%% file: corl_main.tex
\PassOptionsToPackage{numbers,compress}{natbib}

\documentclass{article}
\usepackage{graphicx} 
\usepackage{xspace}
\usepackage{subcaption} 
\usepackage{booktabs}
\usepackage[preprint]{corl_2026}
\hypersetup{bookmarks=true}
\usepackage{times}
\usepackage{xspace}
\usepackage{float}
\usepackage{algorithm}
\usepackage{algpseudocode}
\usepackage{comment}
\usepackage{pgfplots}
\usepackage{eqparbox}
\usepackage{enumitem}
\usepackage{wrapfig}

\usepackage[numbers]{natbib}
\usepackage{multicol}
\usepackage{tcolorbox}
\tcbuselibrary{listings,breakable}
\usepackage{alltt}
\definecolor{ColorBaseline}{RGB}{17,138,178}
\definecolor{ColorBaselineClutter}{RGB}{6,214,160}
\definecolor{ColorVLM}{RGB}{7,59,76}
\definecolor{ColorI1}{RGB}{7,59,76}
\definecolor{ColorI2}{RGB}{17,138,178}
\definecolor{ColorI3}{RGB}{6,214,160}
\definecolor{Fail}{RGB}{255,209,102}

\algnewcommand{\parState}[1]{\State \parbox[t]{\dimexpr\linewidth-\algorithmicindent -4ex}{ \strut#1\strut}}
\algdef{SE}[SUBALG]{Indent}{EndIndent}{}{\algorithmicend\ \textbf{indent}}
\algtext*{Indent}
\algtext*{EndIndent}

\title{LENS: LLM-guided Environment Simplification for Planning and Control in Clutter}

\author{
  Aileen Liao \And
  Rachel Holladay \And
  Dinesh Jayaraman \And
  Michael Posa \\[0.5em]
  University of Pennsylvania \\
  \texttt{\{aliao, rhollada, dineshj, posa\}@seas.upenn.edu}
}

\begin{document}
\maketitle


\newcommand{\methodname}{LENS\xspace}
\begin{abstract}
Despite recent advances in general-purpose robotic manipulation, real-world multi-object clutter remains challenging to handle for today's prevalent approaches. The problem scales in complexity due to more objects and collisions, more unpredictable contact physics, distractors, and task ambiguity. Bridging this gap to real-world deployment requires effective scene abstractions; yet today, producing such abstractions requires extensive task-specific manual engineering, which does not scale. These abstractions are costly to generate and difficult to adjust or fine-tune. We instead propose a plug-and-play fix to automatically generate scene-specific, task-specific, adaptively updating abstractions on top of existing planning and control stacks. \textbf{L}LM-guided \textbf{En}vironment \textbf{S}implification (\methodname) produces a de-cluttered abstracted scene representation by merging (e.g., stacked objects) or pruning (e.g., distant objects) scene entities in a closed loop in response to task progress. These dynamic, task-relevant abstractions are versatile and easy to use. In our experiments, we show that \methodname improves classical planning, model-based control, and a vision-language-action model, across a diverse set of highly cluttered manipulation scenes. Project website: \url{https://lens-2026.github.io/}.
\end{abstract}



\section{Introduction}
A messy home is a happy home, but a robot’s nightmare. As scene clutter grows in unstructured real-world environments, the number of objects, contacts, decision branches, and interaction modes become limiting factors for solvability and real-time performance in planning and model-based control. Learning offers no silver bullet here; learned policies face difficulty in cluttered scenes due to semantic and spatial ambiguity, as well as distribution shifts. So, current research often focuses on curated environments with few, task-relevant objects, bearing little resemblance to the real-world.

This challenge cuts across the full robot control spectrum. Optimization- and planning-based methods must reason over every object in the scene, so their computational cost and failure modes scale with clutter even when most objects are task-irrelevant. Learned controllers, including modern vision-language-action models, face a parallel difficulty: trained on tidy, curated scenes, they encounter severe distribution shift when deployed in dense, unstructured environments, leading to degraded affordance estimates and incorrect contact predictions. In both regimes, performance collapses as scenes grow complex and with no way to focus on what matters. Without a flexible mechanism to suppress task-irrelevant objects before planning or control begins, the combinatorial and distributional burdens imposed by clutter are inherited in full by every downstream method, making the gap between lab and real-world not merely a matter of scale but of fundamental tractability.



\begin{figure}[t]  
\centering
    \includegraphics[width=1.0\linewidth]{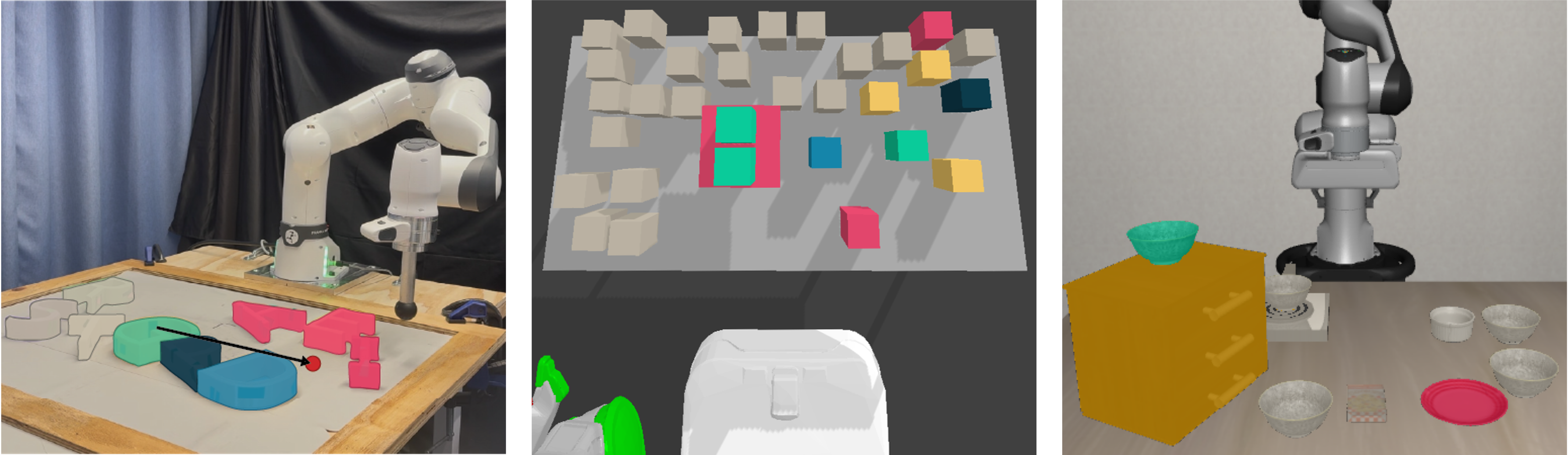}
\caption{
Scene complexity is reduced via pruning and merging.
(Left) The control task is to move the green ``G'' (green) to a target location. \methodname identifies that the ``S'', ``T'', and ``R'' (gray) can be ignored; the ``A'', ``E'', and ``I'' (pink)  are relevant, but can be thought of as a single object (merged); the ``P'' (dark blue) and ``D'' (light blue) are relevant and must be modeled. 
(Middle) for a TAMP problem, the robot must pick and move the three green cubes. \methodname identifies nearby objects as relevant, but prunes the vast majority (gray).
(Right) Using a VLA, the green bowl must be moved to the red plate. Other bowls clutter the scene, but \methodname identifies that they can be pruned from the observation before calling the VLA.
}
\label{fig:fig1}
\vspace{-5mm}
\end{figure}


To address this, we introduce \textbf{L}LM-guided \textbf{En}vironment \textbf{S}implification (\methodname), a general, scalable, task-agnostic method for iteratively determining and updating task relevancy using multimodal large language models (MLLM) with task-success feedback. We consider various kinds of task relevance:
semantic (``what is contextually relevant to the goal?''), geometric (``what blocks a desired motion?''), and dynamic (``what will move or be affected?''). Iterative scene refinement occurs via two elementary operations shown in Figure \ref{fig:fig1}: (i) \emph{pruning}, which removes objects that can be ignored by the planner or controller, and (ii) \emph{merging}, which groups objects that are functionally or dynamically coupled into a single composite entity. The resulting reduced scene representation preserves physical grounding while significantly reducing the combinatorial complexity faced by downstream planners and controllers.

However, zero-shot VLM predictions can suffer from hallucinations or reasoning errors, and the environment state may evolve over time. We therefore implement closed-loop re-prompting triggered by failure: feedback from an infeasible plan or control failure is passed back to the VLM to iteratively correct the abstraction. 



We evaluate \methodname as a plug-and-play component in multiple robotic stacks spanning task and motion planning, model based control, and learned controllers in contact-rich manipulation, demonstrating improved scalability, success rates, and computation time in cluttered environments. The core innovation lies in leveraging modern VLMs to shift the burden of scene abstraction from hand-crafted heuristics to an automated, task-agnostic semantic reasoning loop --- bridging the gap between algorithms that function in structured laboratory settings and the unbounded complexity of the real world. Our contributions are:
\begin{itemize}[leftmargin=*,itemsep=0pt,topsep=0pt]

\item A general, task-agnostic, closed-loop method for estimating task relevance using vision-language models and constructing task-focused scene abstractions. 

\item An integration of this abstraction layer with three distinct state-of-the-art paradigms: (i) Task and Motion Planning (TAMP) \citep{garrett2020pddlstream}, (ii) optimization-based contact-implicit control (C3+) \citep{Aydinoglu2024, bui2025push}, and (iii) the $\pi_0.5$ large-scale Vision-Language-Action (VLA) model  \citep{intelligence2025pi}.

\item An experimental evaluation in cluttered tabletop settings demonstrating improved scalability, robustness, and efficiency with increasing object count.
\end{itemize}

\begin{figure*}[t]
    \centering
  
    \includegraphics[width=0.8\textwidth]{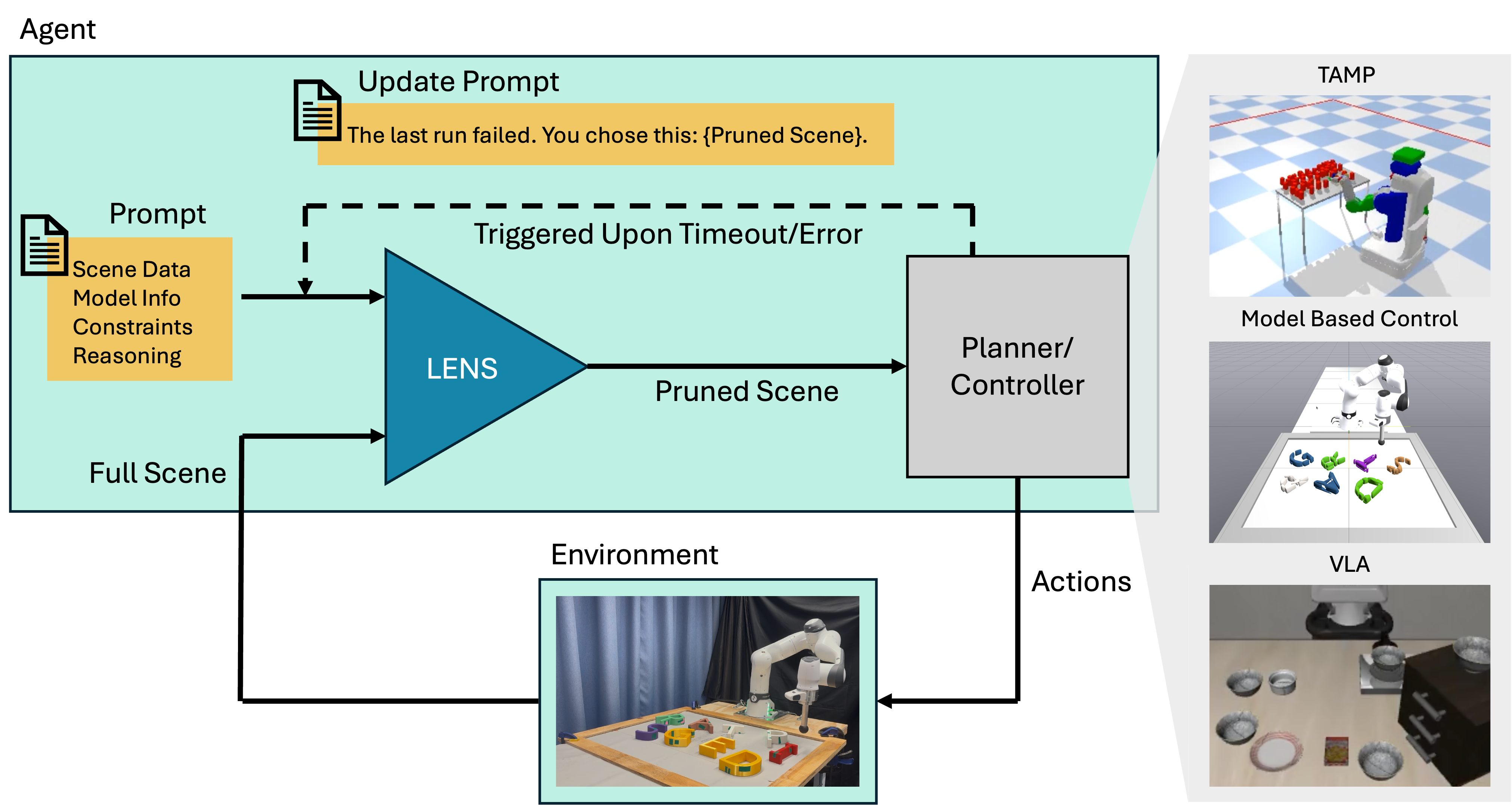}

    \caption{System overview of 
      \methodname. Given a task description via the full scene and prompt information, \methodname (blue) prunes and merges objects to produce an abstracted scene, which is passed to the planner or controller (grey). When triggered, \methodname revises its scene pruning. This approach generalizes to TAMP, model-based controllers, and VLAs.  
    } \label{fig:fig2}
  
  \vspace{-5mm}
\end{figure*}


\section{Related Work}
VLMs have become a versatile tool for enabling robots to reason about their environment, broadly falling into three patterns: (i) generating or sequencing actions and plans, (ii) guiding/constraining planning and search, or (iii) shaping scene abstractions. Our work falls into (iii) and differs from prior approaches by actively reducing the world model before planning or control is performed.

\textbf{VLMs for Action and Plan Generation}. A large body of recent work uses VLMs as high-level planners, program generators, or skill sequencers~\citep{huang2025thinkact, yang2025lohovla, pmlr-v305-feng25b, lin2025onetwovla}. Some methods leverage VLMs to decompose tasks into subgoals and generate executable programs~\citep{saycan2022arxiv, codeaspolicies2022}, while others directly output symbolic plans and action parameters~\citep{wang2024llmˆ}. As these methods operate on a given world representation, they are complementary to our approach of producing that task-relevant representation.

\textbf{VLMs for Guiding and Constraining Planning}. A related line uses VLMs or LLMs to guide classical planners by generating subgoals~\citep{yang2024guidinglonghorizontaskmotion}, predicting constraints~\citep{kumar2024open, lee2024prime}, scoring plans~\citep{huang2022inner}, or inferring feasibility before planning begins~\citep{yan2025using}. In all cases, the underlying planner remains unchanged and the world model is fixed. In contrast, \methodname adapts the scene itself before any planning occurs, operating as a task-agnostic front-end to any downstream system.

\textbf{VLM-Generated Scene Representations and Abstractions}. Most relevant to our approach, several works use vision-language models or learned perception systems to build richer semantic representations of scenes, such as object-centric scene graphs, relational graphs, or affordance-based abstractions~\citep{pmlr-v155-nguyen21b, zhu2021hierarchical, 
silver2021planning, zhang2024learn2decompose, braun2022rhh, qian2024task}. However, these 
approaches typically maintain a complete scene representation, leaving the planner 
or controller to contend with full combinatorial complexity.


A complementary line of work in model-based control addresses complexity through contact-implicit model reduction \cite{bui2024enhancing}, shape approximation for improved contact tractability \citep{nechyporenko2025morphit}, or dynamic-resolution and lumped-object models for piles~\citep{wang2023dynamic, qian2024task}. These approaches, while promising, are ultimately tied to specific tasks or methods.

In visuomotor policies, language-conditioned perception~\citep{shridhar2021cliport} 
acts in clutter without reducing the underlying world model. Observation 
filtering methods~\citep{mirjalili2026augmented, zhang2025peek} do reduce visual 
inputs but require manually specified labels or fine-tuned models and do not extend 
to TAMP or model-based control.


\section{The \methodname Framework}

Our approach constructs task-focused scene abstractions using a VLM, and iteratively refines these abstractions during task execution by monitoring task success. We first describe the core \methodname framework, before discussing its instantiation in downstream systems in the subsequent sections.

The shared core features of \methodname are shown in Figure \ref{fig:fig2} and Algorithm~\ref{alg:controller_vlm}. Given a natural language task description $\tau$ and a scene description $\mathcal{O}$ as input, the VLM (in our implementation, GPT-4o) is prompted to output information to construct a structured scene abstraction $\tilde{\mathcal{O}}$.  
$O$ and $\tilde{\mathcal{O}}$ take a form specific to the particular planning or control framework. For example, in TAMP, $\mathcal{O}$ might consist of object-level features and predicates (e.g., mass, pose, and identity). For a VLA, it is instead an object segmentation mask overlaid on an image observation.

While the full VLM prompt is in Appendix~\ref{appendix:condensedprompt}, we highlight the salient points here. The VLM must provide a list of lists of task-relevant objects critical to task completion to the downstream systems. Sublists are groups of objects that may be merged and treated as a single entity without hurting task performance due to functional or spatial coupling.

\begin{wrapfigure}{r}{0.5\textwidth}
\vspace{-1em}
\begin{minipage}{0.5\textwidth}
\begin{algorithm}[H]
\caption{The \methodname Framework}
\label{alg:controller_vlm}
\begin{algorithmic}[1]
\Require Task $\tau$, objects $\mathcal{O}$, system $\mathcal{S}$, max retries $N$
\Ensure Task execution via abstracted scene $\tilde{\mathcal{O}}$
\For{$i = 1$ to $N$}
    \State \textbf{Generate} $\tilde{\mathcal{O}}$ \textbf{via VLM}
    \Indent
        \State Prompt VLM from $\tau$, $\mathcal{O}$, $\mathcal{S}$; parse output
        \State Prune \& merge objects into $\tilde{\mathcal{O}}$
        \State Run controller on $\tilde{\mathcal{O}}$
    \EndIndent
    \If{execution succeeds}
        \State \Return success
    \Else
        \State Append failure feedback to prompt
    \EndIf
\EndFor
\State \Return failure
\end{algorithmic}
\end{algorithm}
\end{minipage}
\vspace{-1em}
\end{wrapfigure}

Given these VLM outputs, we generate the scene abstraction $\tilde{\mathcal{O}}$ by implementing ``pruning'' and ``merging'' operations on the original scene description $\mathcal{O}$ (Fig.~\ref{fig:fig1}), as follows.

\begin{itemize}[leftmargin=*,itemsep=0pt,topsep=0pt]

\item \textbf{Pruning}. Objects excluded from the list are omitted (``pruned''). The definition of pruning differs for each system and can range from a state space reduction in TAMP to a geometric/collision pairs reduction in C3+. At a high level, this eliminates unnecessary decision variables for TAMP, contact constraints for MPC, and visual distractors for VLA models. 

\item \textbf{Merging}. For each sublist of objects, \methodname merges the rigid bodies into a single abstract body whose collision model is built by aggregating the members’ geometry into one co-moving description that is recomputed from the members’ poses at run time. This reduces the number of bodies and contact pairs while retaining a conservative envelope of the group’s physical extent, depending on the geometry representation used by the downstream system. 
\end{itemize}

An important benefit of \textit{autonomously} constructing the reduced scene representation $\tilde{\mathcal{O}}$ is that it can also be autonomously revised when necessary. In \methodname, feedback is triggered by time-outs or error codes (eg. workspace limits, runtime errors) in downstream systems. Then, a framework-specific feedback prompt along with the previous $\tilde{\mathcal{O}}$ is appended to the existing prompt and current scene, and \methodname re-queries the VLM for a revised scene reduction (See Fig.~\ref{fig:fig2}). In this manner, \methodname injects high-level task semantics into a dynamic scene-abstraction process without needing manually specified task-dependent heuristics.

\section{Implementing \methodname In Various Control Stacks} 
To illustrate its versatility, we now discuss how \methodname-generated scene representations $\tilde{\mathcal{O}}$ integrate into various representative downstream planning/control frameworks (more details in Appendix~\ref{appendix:background}).

\subsection{Task and Motion Planning (TAMP) Stack}
Task and Motion Planning formulates manipulation as a hybrid discrete–continuous problem, jointly solving for the sequence of parameterized actions and the parameters of those actions, which are governed by geometric and physical constraints~\citep{garrett2021integrated}.
Many TAMP frameworks operate on a domain description including a state space defined 
through lifted predicates~\cite{mcdermott1998pddl}, parameterized actions with 
preconditions and effects, and hybrid constraint solvers (motion planners, grasp 
generators, collision checkers).

A strength of TAMP lies in its explicit structure: symbolic reasoning enables planning with correctness guarantees while accounting for geometric and physical constraints. However, TAMP algorithms typically search through the decision tree generated by this structure and are therefore limited to problems of modest scope. While domain-specific relevance heuristics can prune the scene in an attempt to limit complexity, they 
are brittle and do not generalize across task goals/layouts or scene semantics, causing rapid performance degradation in unstructured, cluttered environments. Without a principled, semantic mechanism for reducing the scene to only what is task-relevant before planning begins, TAMP cannot be viably deployed outside of curated, low-clutter settings. An anecdotal example of VLM-based pruning vs a geometric heuristic is shown in Appendix~\ref{appendix:tamp-anecdote}.

\textbf{\methodname Implementation.} $\mathcal{O}$ is a graph of objects with poses, masses, and relations. $\tilde{\mathcal{O}}$ defines which objects are treated as active decision variables in PDDLStream~\citep{garrett2020pddlstream}. Pruned objects participate in collision checking but have fixed poses and cannot be acted upon. Merged groups that must be stably movable together (e.g., a stack of objects that can be moved via the base) adopt a unified ID with the lowest supporting object as parent, and cannot be separated during planning. Feedback triggered at 120s prompts expansion of the object selection.

\subsection{Contact-Implicit Model Predictive Control Stack}
Contact-implicit trajectory optimization and MPC formulate manipulation as a continuous optimization problem that jointly reasons over robot motion, object dynamics, and contact interactions~\citep{Aydinoglu2024}. This enables controllers to reason about making/breaking contact, pushing multiple objects, and exploiting incidental interactions. However, each  object introduces new decision variables, collision constraints, complementarity conditions, and contact pairs. The resulting problems are nonconvex and scale poorly with object count.


In cluttered scenes, this creates a fundamental barrier to real-world deployment. Controllers that simplify contact modeling, whether by limiting contact pairs~\cite{bui2024enhancing} or softening constraints~\cite{le2024fast}, reduce computational burden at the cost of physical fidelity, producing plans that fail on contact-rich tasks. Alternatively, modeling all interactions becomes intractable as object count grows. Neither path scales: the first sacrifices correctness, the second sacrifices tractability. A principled front-end that reduces the scene to its task-relevant structure is a prerequisite for deployment outside the lab.

\textbf{\methodname Implementation.} The scene $\mathcal{O}$ is a set of objects with poses and names. Unlike TAMP, merged groups need not be physically stable;    objects are clustered if their contact interactions can be reasoned about collectively (e.g., a pile to be pushed aside). The merged body's geometry is updated in real time as the scene evolves, though the controller does not model internal shape changes. Pruning reduces the number of bodies, contact pairs, and linear complementarity problem dimensionality. Feedback is triggered at 250 control iterations or on an error code.

\subsection{Vision-Language-Action Model Stack}
Vision-Language-Action models combine visual encoders and language understanding modules to predict actions from perceptual observations and task descriptions~\citep{intelligence2025pi}. These models excel at capturing semantic relationships, object affordances, and task intent, and can generalize to novel objects and instructions without explicit symbolic modeling.

However, VLAs typically operate over dense, unstructured perceptual representations. Scene understanding is implicit and distributed across learned feature spaces. So, VLAs lack explicit mechanisms for enforcing physical constraints, reasoning long-horizon, or guaranteeing task feasibility. In cluttered environments, these limitations compound: irrelevant objects, occlusions, and novel contact configurations introduce distribution shifts that corrupt affordance predictions and produce physically implausible actions. 

\textbf{\methodname Implementation.} $\mathcal{O}$ is an overhead image with object bounding boxes. Merging is not used; $\tilde{\mathcal{O}}$ is a flat list of relevant objects. Excluded objects are inpainted (See Fig.~\ref{fig:vla-hardware}), producing a filtered image passed to $\pi$-0.5~\citep{intelligence2025pi}. This reduces visual clutter and suppresses distractor affordances without modifying the policy. Given the short task horizon, no feedback re-queries are needed.


\section{Experiments and Results}


We evaluate \methodname across tabletop manipulation tasks against baselines 
on unfiltered observations. VLM query time averaged 1.76s (model-based control, comparable times for TAMP and VLA), which is negligible relative to execution time and excluded from runtime comparisons. \methodname improves success rates for all three downstream modules. In TAMP, scene pruning and object merging reduce distractor interference and enable reliable plan search under heavy clutter (Fig~\ref{fig:tamp-results}). In model-based control, LLM-guided filtering maintains stable execution times as scene complexity scales, while the baseline degrades by orders of magnitude (Fig~\ref{fig:ablation}). For VLA policies, task-focused abstraction recovers substantial performance lost to visual clutter in both simulation and hardware (Table~\ref{tab:vla_results}). These results demonstrate that a single abstraction framework generalizes across planners, and model-based and learned controllers. We analyze each set of results in the following sections.

\subsection{Task and Motion Planning}

\begin{figure}
    \centering
    \includegraphics[width=1.0\linewidth]{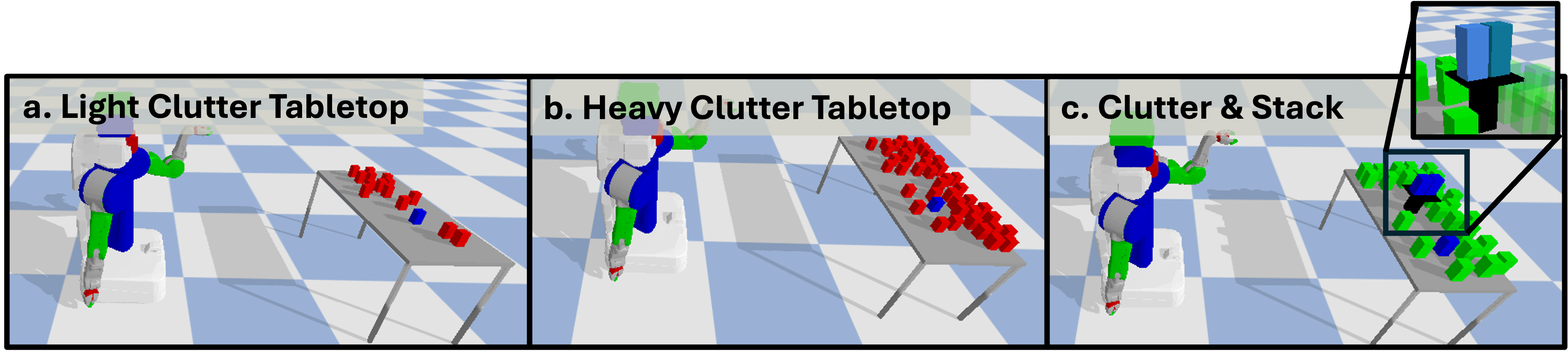}
    \caption{\methodname was evaluated on three TAMP environments: (a) a lightly cluttered tabletop, (b) a heavily cluttered tabletop, and (c) a cluttered tabletop with a stack. The stack consists of a tray (black) with two goal objects (blue). In each environment, a goal object is placed on the tabletop.}
    \label{fig:tamp-envs}
    \vspace{-14pt}
\end{figure}

\begin{figure}[b]
    \centering
    \begin{subfigure}[t]{0.5\linewidth}
        \centering
        \vbox to 3.0cm{\vfil\input{figs/tamp_countSuccess}\vfil}
        \caption{In each bar in the top row, Iteration 1 and 2 success rates are stacked. Baseline is rerun twice for fairness.}
        \label{fig:tamp-results}
    \end{subfigure}
    \hfill
    \begin{subfigure}[t]{0.49\linewidth}
        \centering
        \vbox to 3.0cm{\vfil\includegraphics[width=7cm, keepaspectratio]{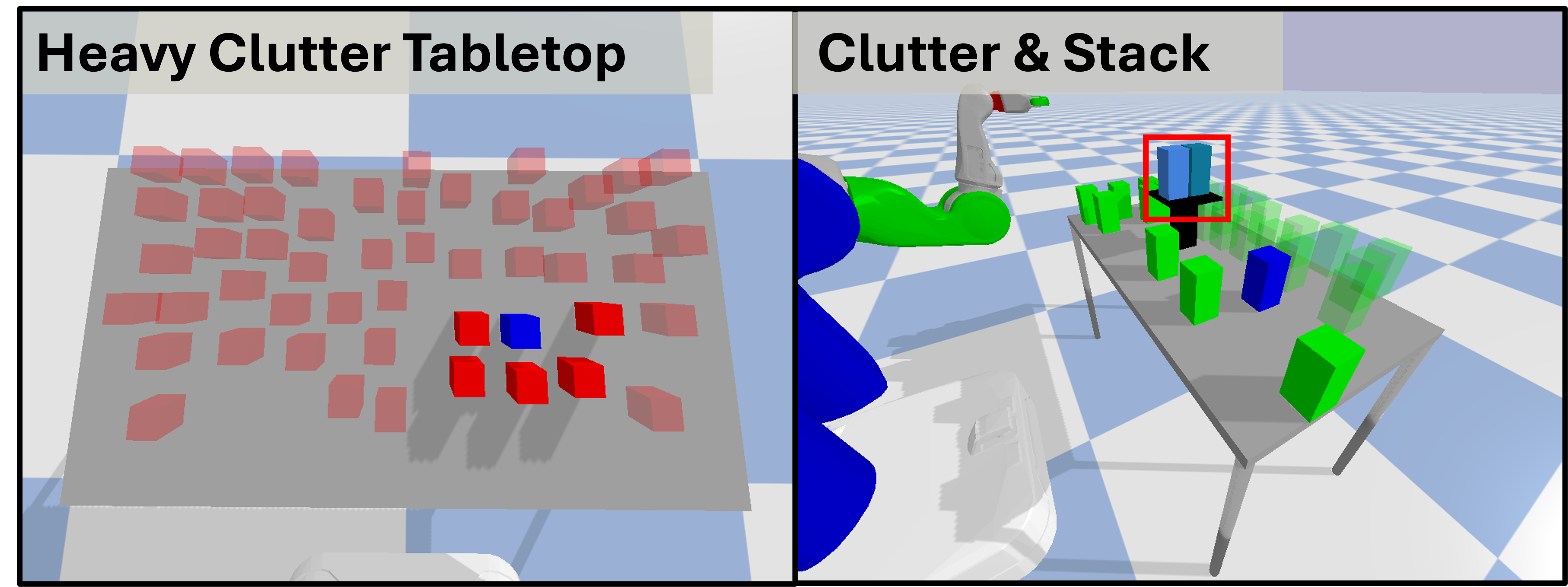}\vfil}
        \caption{Goal objects are blue. Semi-transparent objects are pruned. The stack is merged into one entity.}
        \label{fig:tamp-prune}
    \end{subfigure}
    \caption{TAMP Results. (a) shows Baseline-TAMP and \methodname-TAMP success rates in three environments arranged left to right in order of complexity: Light Clutter, Heavy Clutter, and Clutter\&Stack. (b) Examples of scene pruning are shown for Heavy Clutter and Clutter\&Stack.}
    \vspace{-15pt }
\end{figure}

To evaluate \methodname's effect of a TAMP framework's performance, we construct three environments with increasing levels of visual and semantic complexity (Figure~\ref{fig:tamp-envs}): (1) light-clutter tabletop, (2) heavy-clutter tabletop, and (3) 
clutter \& stack scenarios.The blue object(s) need to be moved to the green goal region; the red objects can be moved if necessary.
Clutter \& Stack includes three target objects; two are placed on a tray, and one is placed directly on a table surrounded by clutter.

Across all environments, we compare \methodname against a baseline operating on unfiltered visual observations. Performance is measured using task success rate over 50 evaluation episodes, as shown in Fig~\ref{fig:tamp-results}. Two iterations of feedback are allowed. Success rates for each iteration are shown stacked.

In light clutter, both methods achieve comparable success; abstraction 
provides limited benefit when the full scene remains tractable. More noticeably in heavy clutter, \methodname-TAMP outperforms the baseline, which frequently times out enumerating distractors. LLM/VLM-guided filtering suppresses irrelevant objects, improving success rates (Fig~\ref{fig:tamp-prune}). In Clutter \& Stack, \methodname again outperforms the baseline, which fails 
due to distractor interference. Merging the stack into a single composite simplifies 
the scene and enables more reliable plan search.




\subsection{Model-Based Control}\label{sec:exp-c3}

We study a planar pushing task (Fig~\ref{fig:push-anything}) inspired by Push Anything~\cite{bui2025push}. 
The goal is for the robot's end-effector to push a goal object (``G'') to a target pose (transparent ``G''). This requires reasoning over object-object, object-robot, and object-ground contact interactions. The scene is cluttered with additional objects compared to Push Anything~\cite{bui2025push}, which was limited to four objects at a time.

\begin{wrapfigure}{r}
{0.5\linewidth}
        \centering
        \vspace{-10pt}
    \input{figs/pruningObjectsAblation}
    \vspace{-15pt}

    \caption{Object Count Ablation. As the object count increases, the baseline execution time increases drastically. \methodname remains constant. }

    \label{fig:ablation}
    \vspace{-15pt}

\end{wrapfigure}

\noindent \textbf{Simulation Experiments:}
We varied the number of objects present in the environment and compared against the baseline controller C3+ that operates over the full scene. Baseline-C3+ does not perform any relevance estimation and reasons over all objects uniformly. Figure \ref{fig:ablation} reports average execution time vs clutter count. Results for each object are averaged across five trials. 
\methodname-C3+ and baseline succeeded on 39/45 and 17/30 trials respectively. 

For small scenes (2–4 objects), the baseline and \methodname exhibit comparable performance. In this regime, the abstraction overhead provides limited benefit as the full scene remains tractable for the underlying planner and controller.
As object count increases, however, baseline-C3+ sharply degrades in performance. At 6 objects, baseline-C3+ becomes prohibitively slow, exceeding \methodname by an order of magnitude ($\sim$1000 seconds). At 7 objects, it reaches over 4000 seconds. In contrast, \methodname-C3+ maintains stable performance across all tested clutter levels, with average execution time of $\sim$40–135 seconds. 
This indicates that the effective problem size is determined by task relevance rather than raw scene complexity. We also found \methodname had higher success rates compared to various distance-based geometric pruning baselines (Appendix~\ref{appendix:simulation-baselines}).


\begin{figure}[t]
    \centering
    \input{figs/hardware}
    \vspace{-2mm}
    \caption{\methodname's task outcomes are shown for Merging and Pruning experiments on hardware. Task successes are recorded for across iterations of VLM requeries.}
    \label{fig:push-anything-real}
    \vspace{-5mm}
\end{figure}

\begin{figure}[b]
    \centering
    \begin{subfigure}[t]{0.48\linewidth}
        \centering
        \vbox to 2cm{\vfil\includegraphics[width=\linewidth]{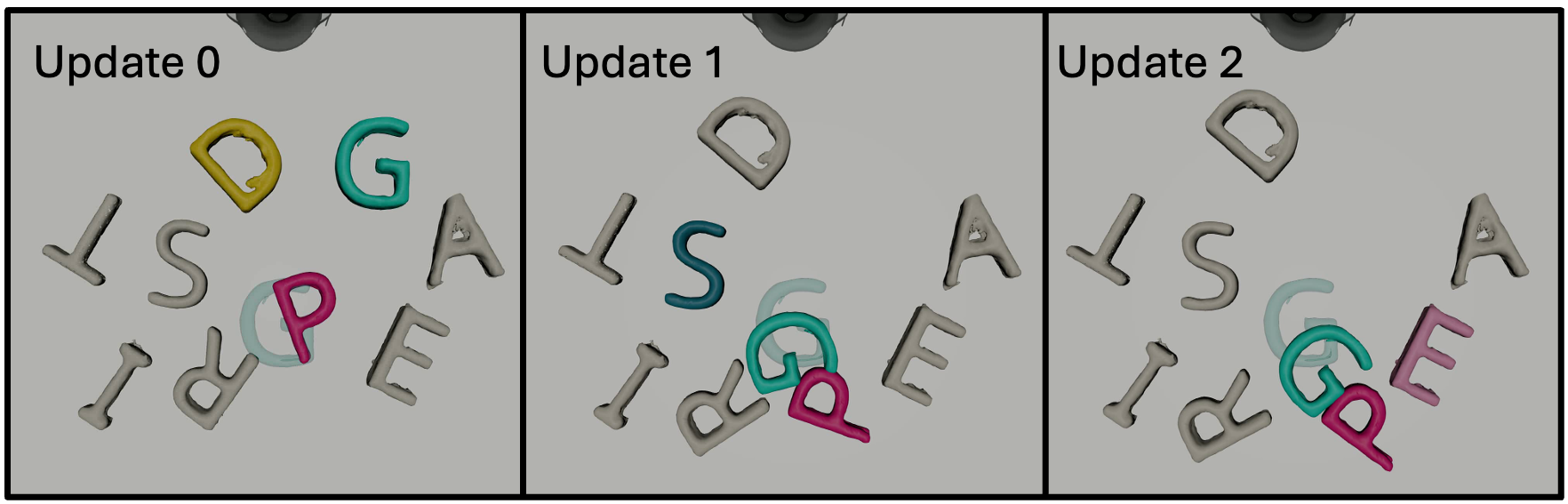}\vfil}
        \caption{Pruned objects are shown in gray. Successive iterations refine the abstraction to adapt to failures, mistakes, or evolving environment states.}
        \label{fig:push-anything-prune}
    \end{subfigure}
    \hfill
    \begin{subfigure}[t]{0.48\linewidth}
        \centering
        \vbox to 2cm{\vfil\includegraphics[width=\linewidth]{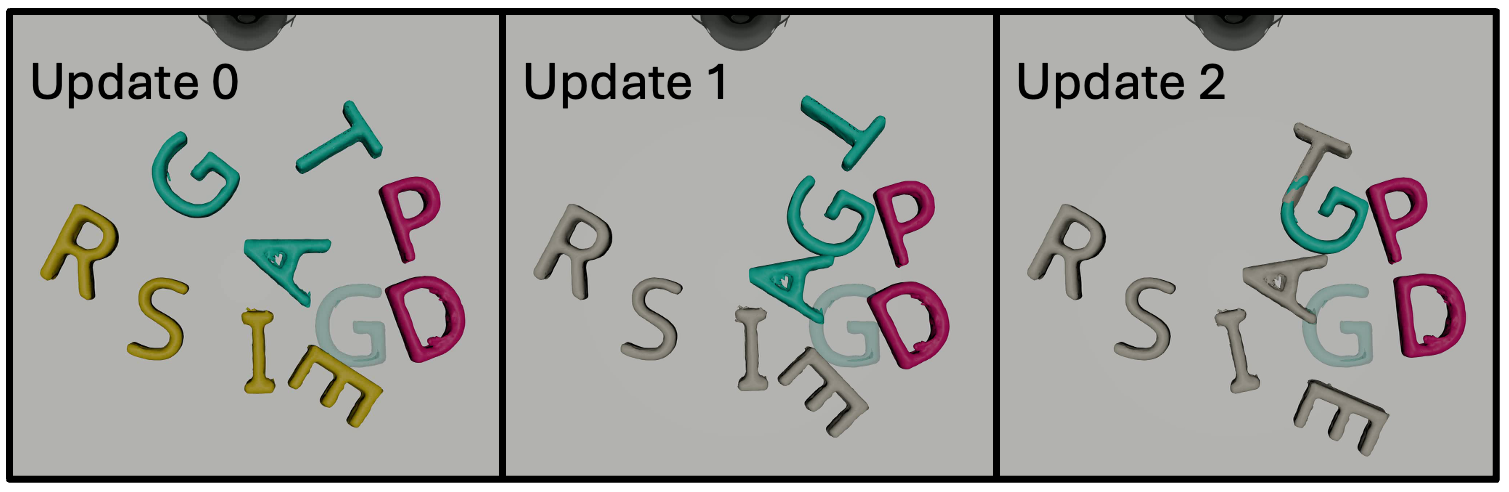}\vfil}
        \caption{Merged groups, shown with uniform colors, become more focused over iterations.}
        \label{fig:push-anything-merge}
    \end{subfigure}
    \caption{Three iterations of (a) pruned and (b) merged scene graph at various times during an episode of multi-object planar pushing (Sec~\ref{sec:exp-c3}). Goal location for target "G" is overlayed.}
    \label{fig:push-anything}
    \vspace{-10pt}
\end{figure}

\noindent \textbf{Real Robot Experiments:} We evaluate two hardware variations: pruning only (precise pushing goal object "G" to a target pose and orientation) and pruning+merging (gross pushing through clutter with a looser 
position-only threshold). Fig~\ref{fig:push-anything-real} shows the success rate broken down by iteration, limited to three feedback loops. For both pruning and merging, the cumulative success rate is high at 80\%, with successes distributed throughout all three iterations rather than concentrated in the first. This suggests that aggressive initial pruning may omit relevant contacts, with subsequent VLM feedback correcting for this and enabling task completion. Failure cases include timeouts and workspace limits. A visualization of three feedback loops on hardware is shown in Fig~\ref{fig:push-anything}. These results demonstrate that iterative VLM feedback enables recovery from execution errors while maintaining high success on hardware in model-based control.




\begin{figure}[t]
    \centering
    \includegraphics[width=1.0\linewidth]{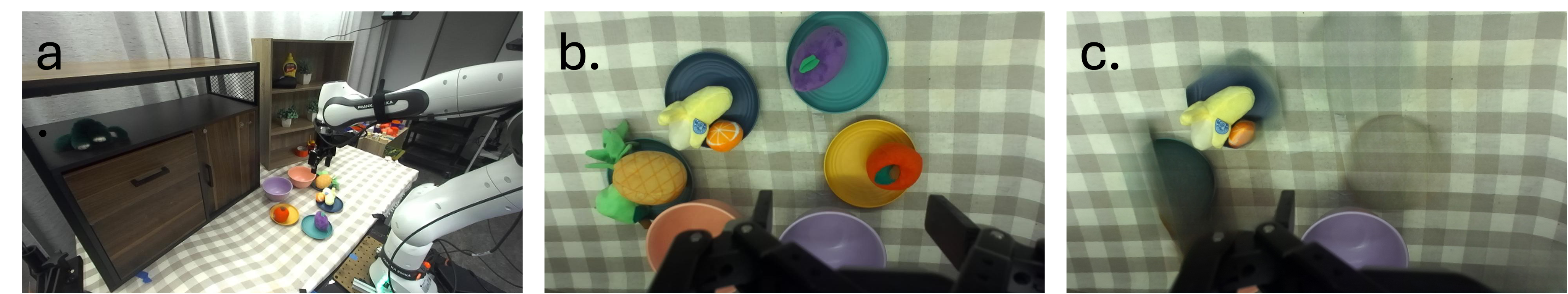}
    \caption{Real hardware VLA experiment. \methodname is applied directly on hand image (b) without simulator bounding boxes. Foundation models ~\cite{liu2024grounding, kirillov2023segany, suvorov2021resolution}  are used to segment and inpaint scenes (c). Performance improvements persist, supporting the validity of the simulation comparison.}
\label{fig:vla-hardware}
\vspace{-10pt}

\end{figure}

\subsection{Vision Language Action Models}

\noindent \textbf{Simulation Experiments:} We next evaluate task-focused scene abstraction with a Vision–Language Action (VLA) using the Spatial LIBERO benchmark with a $\pi_{0.5}$ policy. Spatial LIBERO consists of tabletop manipulation tasks that require reasoning over relative spatial relationships, such as placing, stacking, or aligning a single object type (bowl) in varying positions across different scenes. To stress test semantic and spatial reasoning, we introduce additional instances of the goal object beyond the original Spatial LIBERO scenes, increasing visual ambiguity and adding distractions. An example scene and VLM pruning selection is included in Appendix~\ref{appendix:vla-sim}. Ground-truth segmentation is used in simulation as open-vocabulary detectors 
perform unreliably under domain mismatch~\citep{liu2024grounding}.

Task success rates averaged over 100 evaluation episodes across 10 tasks (Appendix~\ref{appendix:vla-sim}) show that clutter significantly degrades VLA performance, with success dropping from 0.85 to 0.5, in keeping with prior reports~\cite{rasouli2025distracted,fei2025libero,zhang2025vla}. \methodname  recovers performance substantially, improving success to 0.69.
The remaining gap highlights limitations of imperfect abstraction and the underlying VLA’s robustness to distribution shift. Overall, these results indicate that task-focused scene abstraction is an effective, lightweight mechanism for improving VLA performance in visually cluttered spaces. 




\noindent \textbf{Real Robot Experiments:}
On hardware, experiments (Figure~\ref{fig:vla-hardware}, 
Table~\ref{tab:vla_results}) do not have privileged bounding boxes. An object detection and inpainting pipeline (Appendices~\ref{appendix:vla-sim},~\ref{appendix:condensedprompt}) is used. $\pi_{0.5}$ picks up one of six fruit plushies and place it in a target bowl. Performance gains persist, 
confirming they reflect scene abstraction.


\begin{table}[h]
\centering
\setlength{\tabcolsep}{8pt}
\renewcommand{\arraystretch}{1.2}
\begin{tabular}{lcccc}
\hline
Method & \multicolumn{4}{c}{Success Rates} \\
       & Pineapple & Pear & Banana & Apple \\
\hline
LENS in Clutter    & 0.5 & 0.3 & 0.3 & 0.7 \\
Baseline in Clutter & 0.2 & 0.0 & 0.0 & 0.0 \\
\hline
\end{tabular}
\caption{Hardware VLA ($\pi_{0.5}$) results (n=10 per fruit). \methodname consistently improves over the baseline across tasks.}
\label{tab:vla_results}
\vspace{-20pt}
\end{table}

\section{Conclusion}
\label{sec:conclusion}
We introduced \methodname, a general, task-agnostic framework that constructs compact, physically grounded scene abstractions by pruning irrelevant objects and merging functionally coupled ones using VLMs and task feedback. Integrated as a front-end to existing pipelines, \methodname improves scalability, robustness, and efficiency across contact-rich, cluttered manipulation scenarios.

\textbf{Limitations.} While abstract representations could be validated in simulation before execution, it is computationally prohibitive for some controllers. The feedback mechanism relies on high-level failure signals rather than explicit causes, and \methodname does not learn from experience across runs.

More broadly, this work argues that as robots are deployed in increasingly complex and unstructured environments, the ability to dynamically construct minimal, task-conditioned world models will be critical for scaling both model-based and learning-based methods. Future work can extend these ideas to richer abstraction operations (eg. via RL, finetuning), tighter perceptual integration, and long-horizon multi-stage tasks where the notion of relevance evolves over time.

\acknowledgments{This work was supported by DARPA TIAMAT under Grant No.HR0011249042, NSF CAREER Awards under Grant Nos. 2238480 and 2239301, the Office of Naval Research (ONR) under Grant No. N00014-22-1-2677, and NSF Science of Learning and Embedded Systems (SLES) program under Grant No. 2331783.}


\clearpage


\bibliography{references}  

\newpage

\appendix
\section{Background Details}
\label{appendix:background}

The abstraction formatting and rules vary across integrations to match the physical and representational requirements of each downstream system. The core \methodname pipeline is unchanged; these differences reflect how $\tilde{\mathcal{O}}$ is expressed in a form each module can consume.

\subsection{\methodname For Task and Motion Planning (TAMP)}
We represent the environment ${\mathcal{O}}$ as a scene graph of objects with known poses, masses, relations (such as stacked, on) and properties (such as graspable). $\tilde{\mathcal{O}}$ is the subset of ${\mathcal{O}}$ whose poses are treated as decision variables in the planning problem if PDDLStream \citep{garrett2020pddlstream}, allowing actions that explicitly modify their pose (e.g., pick, place, or push). In contrast, static objects excluded from ${\mathcal{O}}$ have fixed poses and cannot be acted upon, though they still impact feasibility.

To determine $\tilde{\mathcal{O}}$, \methodname is provided with a list of object IDs and their associated mass and pose, along with a goal specification represented as a list of tuples $(o_i, g_i)$, where each tuple corresponds to an object and its desired goal pose.

Sublist groups must be stably movable together (eg. stacks). They adopt a new, unified ID, and the parent object is the lowest supporting object in the stack. Sublist groups have their geometries fused and maintain the mass of the base object. A set of merged objects cannot be moved separately.  
~
To maintain feasibility constraints, collision checking over the full scene $\tilde{\mathcal{O}}$ is still required.

In \methodname, feedback is triggered when no solution has been found or when the maximum timeout of 120 seconds has been reached. 
The feedback includes the previous scene representation and a description of the failure in the form of: "The last run failed. You chose [previous representation]. Choose a larger set of objects from the scene that is not just goal objects." This feedback encourages larger selection sets to expand the plan search. The feedback is appended to the original prompt and new scene representation.

\subsection{\methodname for Model-Based Control}

We represent the scene as a set of objects $\mathcal{O}$. Given the set of object's poses and names, \methodname outputs $\tilde{\mathcal{O}}$, the reduced set of objects. 

Note that sublist sets are not required to be stably movable and are instead clustered if their contact sampling can be reasoned about together. For example, piles of objects are not physically stable but can be pushed aside together. 
The collective union shape of the sublist sets are updated in realtime as the scene changes.
Since the controller treats the pile as a single object, it does not model or predict how its shape will shift over time.

We integrate this task-focused scene abstraction $\tilde{\mathcal{O}}$  into the C3+ contact-implicit controller \cite{bui2025push} that models manipulation as a linear complementarity system (LCS) and solves a linear complementarity problem (LCP) at each control step. Selected objects contribute a rigid body to the plant, potential contact pairs, and sampled surface regions. Pruning reduces the number of bodies, contact pairs, and LCP dimensionality, including the surface sampling space. Both contact generation and contact resolution therefore operate on a reduced-order system.

Feedback is triggered when 250 controller loop iterations was reached or an error code was received. The feedback includes the previous scene representation and a description of the failure in the form of: ”Your previous answer was not solvable. You selected [past object selection]." The feedback is appended to the original prompt and new scene representation.

\subsection{\methodname For Vision Language Action Model}

We integrate task-focused scene abstraction into a vision-language-action (VLA) policy by filtering visual inputs using the VLM output. Before querying the VLA ($\pi$-0.5 trained on the LIBERO-Spatial dataset \citep{liu2023libero}), the VLM is provided with ${\mathcal{O}}$, an overhead scene image and ground-truth object bounding boxes from the simulator, a semantic goal sentence, and the original prompt. ~${\tilde{\mathcal{O}}}$ does not include sublists for the VLA.

To construct an input for the VLA using ${\tilde{\mathcal{O}}}$, a new image is created with only the objects in ${\tilde{\mathcal{O}}}$ included. This is implemented via generative-modeling based image inpainting, yielding less cluttered, task-focused observations. Details are found in Appendix~\ref{appendix:vla-sim} The resulting filtered images are then used as input to the VLA policy.
This preprocessing step reduces visual clutter and semantic ambiguity by suppressing distractor objects unrelated to task execution. As a result, the VLA receives observations that are more closely aligned with the task-relevant scene structure encountered during training, improving robustness in cluttered environments. Beyond the training distribution, this filtering reduces the number of visually plausible affordances in the scene, simplifying action selection under clutter. 
Given the short task horizon, the filtered scene remains valid throughout execution, so no additional VLM re-queries are needed.

\section{TAMP Simulation Examples}
\label{appendix:tamp-anecdote}
\begin{figure}[H]
    \vspace{-10pt}
     \includegraphics[width=1.0\linewidth]{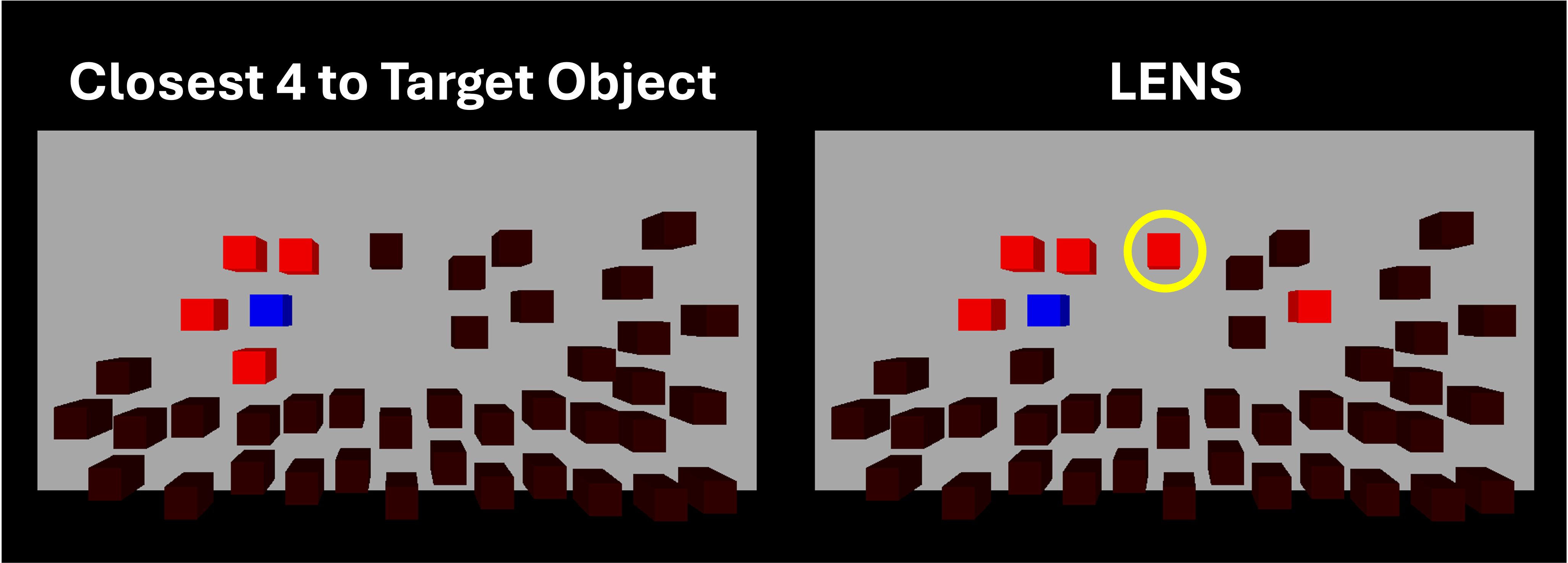}
    \caption{Example TAMP scene where a distance-based baseline fails. The set of nearest objects to the goal (highlighted) fails to include a critical blocking object further away (circled yellow), which must be removed to enable a successful approach to the target. \methodname identifies the blocker as task-relevant.}
    \label{fig:tamp_anecdote}

\end{figure}

An anecdotal example (Figure \ref{fig:tamp_anecdote}) showcases an iteration of \methodname pruning compared to a distance based metric such as only planning over the closest 4 target objects. While \methodname may not be perfect and my have spurious selections, it is able to maintain the critical objects when multiple blocking objects need to be removed for access. 

\section{C3+ Simulation Baseline Comparisons}
\label{appendix:simulation-baselines}
\begin{figure}[H]

    \centering
     \input{figs/baselines}
    \caption{C3+ distance-based pruning baseline (n=30). Nearest-3 objects or nearest objects in $r$ radius are passed to C3+. \methodname outperforms the heuristics by capturing task-relevant objects regardless of proximity.}
    \label{fig:c3}
    \vspace{-10pt}

\end{figure}
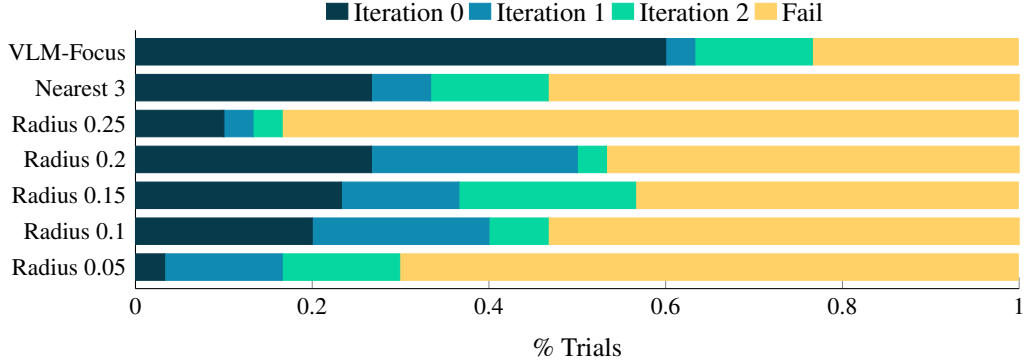

We compared \methodname to distance-based geometric pruning baselines for C3+ (Figure~\ref{fig:c3}). These baselines select the 3 spatially nearest objects or the objects within $r$m radius to the goal and pass them to the downstream system. Results show that geometric proximity is a reasonable heuristic, but task success may require more distant objects  for multi-step planning.
This supports the claim that the VLM's task-relevant reasoning adds value beyond simple spatial heuristics, which typically require human-engineering and fine-tuning. 

\section{VLA Experiments}
\label{appendix:vla-sim}

In simulation, object masking is implemented using ground-truth segmentation rather than learned detectors. This choice is motivated by the observation that open-vocabulary detectors such as GroundingDINO ~\citep{liu2024grounding} perform unreliably in simulated scenes due to domain mismatch, leading to degraded masking quality unrelated to the abstraction mechanism itself. 

In hardware, \methodname selections are passed through GroundingDINO~\cite{liu2024grounding}, 
SegmentAnything~\cite{kirillov2023segany}, and LaMa~\citep{suvorov2021resolution} 
to inpaint over task-irrelevant object masks. A hardware-specific version of the VLM prompt (see Appendix~\ref{appendix:condensedprompt}) was used to produce outputs compatible with the GroundingDINO detection pipeline.

\begin{figure}[h]
    \centering
    \begin{subfigure}[t]{0.48\linewidth}
        \centering
        \vbox to 5cm{\vfil\includegraphics[width=\linewidth]{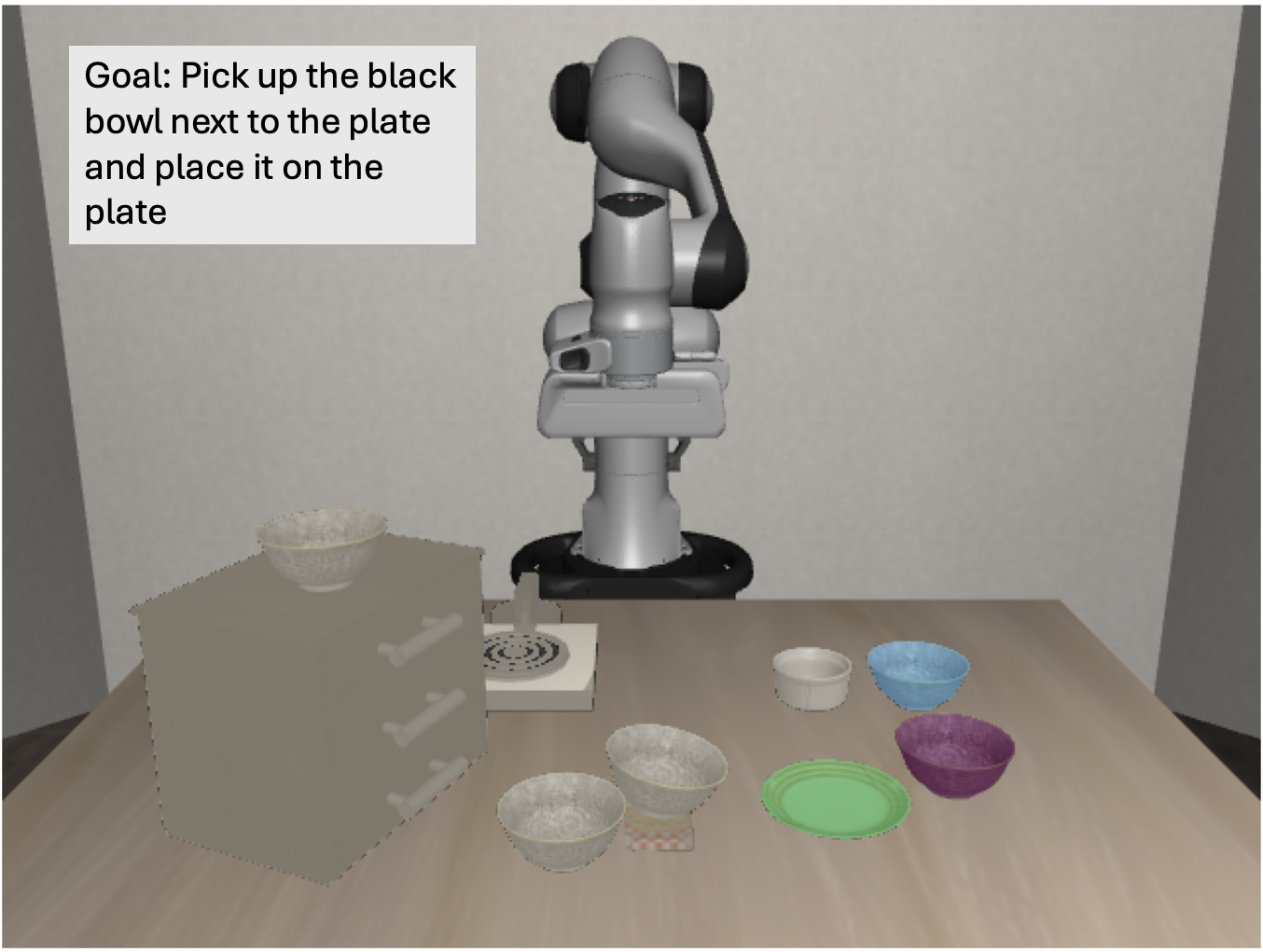}\vfil}
        \caption{VLA Pruning Visualization. Grey objects in the scene have been pruned away, while objects with colored mask overlays are identified as semantically and spatially goal relevant. The goal (masked green) and target objects (masked purple) are selected. In addition, a second distractor goal is selected due to ambiguity in the prompt.}
        \label{fig:vla-pruning-vis}
    \end{subfigure}
    \hfill
    \begin{subfigure}[t]{0.48\linewidth}
        \centering
        \vbox to 5cm{\vfil\input{figs/vlaSuccess}\vfil}
        \caption{The VLA is executed directly on the cluttered scene in Baseline in Clutter. In LENS, task-irrelevant objects are removed using our VLM-based scene abstraction prior to querying the VLA. Baseline without Clutter corresponds to the original, uncluttered LIBERO setup.}
        \label{fig:vla}
    \end{subfigure}
\end{figure}


\vfill

\section{Prompts}
\label{appendix:condensedprompt}
\vspace*{0pt}  

\begin{tcolorbox}[title=Condensed VLM Prompt, colback=gray!5, colframe=gray!70]
You are a precise symbolic robot planner assistant. Your task is to categorize objects into lists based on physical properties and spatial relations.\\

\#\#\# SCENE DATA\\
- Environment: You are operating in a cluttered tabletop scene. \\
- Goal: \{goal\} \\
- Objects: \{objects\}\\
- Stacked: \{stacked\}\\
\\
\#\#\# MODEL INFORMATION\\
You are filtering the scene for a TAMP planner. This is a tabletop scene with multiple goal objects. \\
List [2] object groupings MUST be stably movable together (cannot just be a cluster).\\

\#\#\# RULES\\
1. Include ALL goal objects and any clutter obstructing or near them. Do NOT omit objects between robot and goal.\\
2. Each sublist groups objects that can be stably moved together (not just spatial clusters).\\
3. Objects sharing the same base must be combined into one sublist — base object listed first.\\
4. Standalone relevant objects with no group are listed as singletons: [obj].\\
5. Order sublists by task importance.\\

\#\#\# REASONING STEPS\\
1. **Goals:** Which objects are part of '{goal}'?\\
2. **Obstructions:** Which objects are stacked on, supporting, or blocking a goal object?\\
3. **Grouping:** Which objects can move together? Merge any that share a base.\\
4. **Order:** Base first within each sublist. Most important sublist first overall.\\

\#\#\# OUTPUT FORMAT\\
Reasoning: one line, 30 tokens \\
Result: [[base, obj, ...], [obj], [base, obj, ...], ...]\\

\#\#\# FINAL OUTPUT\\
Reasoning:\\
Result:
\end{tcolorbox}  


\begin{tcolorbox}[title=TAMP Cluttered Tabletop, colback=gray!5, colframe=gray!70, breakable]
    You are filtering the scene for a TAMP planner. The robot is a PR2 movable base robot located at pos = [2.5, 0, 0]. \\
    This is a cluttered tabletop scene with one goal object. You should reduce clutter to reason over.
    Think about what would obstruct robot from goal object.
  \end{tcolorbox}  

\begin{tcolorbox}[title=TAMP Clutter \& Stack, colback=gray!5, colframe=gray!70, breakable]
    Filtering a cluttered tabletop scene for a TAMP planner. Robot is a PR2 at pos = [2.5, 0, 0].\\
    There is clutter around EACH goal object and on the table.
    DO NOT FORGET THERE IS BOTH TRAY GOAL OBJECTS AND TABLE GOAL OBJECTS. You need to list clutter around all beyond just goal objects!
    Clutter can be singletons.
\end{tcolorbox}  
  
\begin{tcolorbox}[title=C3+ Pruning, colback=gray!5, colframe=gray!70, breakable]

    You are a model based controller. You are trying to reduce number of objects to reason over to reduce linear complimentarity problem size and contact constraints.
    Mergings in this model are object groupings (excluding the goal object) to be reasoned about together for computatinal efficiency. They are encouarged.
    CHOOSE AT MOST 4 objects! \\
    Narrow your selection to less objects when closer to the goal

  \end{tcolorbox}

\begin{tcolorbox}[title=C3+ Merging, colback=gray!5, colframe=gray!70, breakable]
    You are a model-based controller optimizing a Linear Complementarity Problem (LCP). Computational cost scales directly with the number of sublists.
    Mergings in this model are object groupings to be reasoned about together for computational efficiency. They are encouraged.\\
    IMPORTANT — THIS SECTION SUPERSEDES RULES AND REASONING STEPS BELOW.\\
    DO NOT LIMIT MERGING to physically stacked or physically grouped objects, any clutter can be merged.
    Ignore base sharing requirements. \\
    Merging means placing multiple object names inside the same sublist: [obj a, obj b, obj c]. This is one sublist, not three. A result with 9 singletons has 9 sublists — that is the WORST possible outcome. \\
    BAD: [[G], [A], [B], [C], [D], [E], [F], [H], [I]] — 9 sublists, maximum LCP cost\\
    GOOD: [[G], [A, B, C, D], [E, F, H, I]] — 3 sublists, minimum LCP cost\\
    Completely irrelevant non-goal objects can be entirely ignored too and not listed as merges or singletons.
    Keep goal object as separate sublist. \\
    MERGES ARE STRONGLY ENCOURAGED FOR NON GOAL OBJECTS!
    When goal object is close to target, singletons are preferred.\\
    AT MOST 4 SUBLISTS TOTAL!!!!!! DO NOT GIVE MORE THAN 4 SUBLISTS!!!!
  \end{tcolorbox}

\begin{tcolorbox}[title=Libero Spatial, colback=gray!5, colframe=gray!70, breakable]
    You are filtering a cluttered scene for a pi0.5 VLA. You are trying to SEMANTICALLY/SPATIALLY SIMPLIFY THE SCENE. Focus on what goal semantics say and spatial relevancy of objects. Remove objects that would confuse the goal!
    Keep in distribution with trained Libero-10 dataset. Keep target location and specific goal object! There is one target object.
  \end{tcolorbox}  

\newpage

\begin{tcolorbox}[title=VLA Hardware, colback=gray!5, colframe=gray!70]

You are helping a robot manipulation pipeline clean up camera images.\\

The robot's current task is: "\{goal\}"\\

Look at the robot workspace image(s) and identify:\\
1. Objects RELEVANT to the task (to be PROTECTED from removal)\\
2. DISTRACTOR objects on the workspace (to be REMOVED from the image)\\

Rules for RELEVANT objects (protect):\\
- ALWAYS include: the robot arm, gripper, and end effector (if visible).\\
  The gripper/end effector is the most critical — it includes the finger pads, gripper jaws, and anything the robot is currently grasping. NEVER remove it.\\
- ALWAYS include: any object the robot needs to interact with for the task\\
- ALWAYS include: any container/bowl/plate the target object is sitting in or on\\
- ALWAYS include: any goal location (basket, bin, tray, bowl) mentioned in the task\\
- DO NOT include: background furniture (table, shelf, wall, floor)\\

Rules for DISTRACTOR objects (remove):\\
- List EVERY single visible movable object that is NOT the task target or robot\\
- Include each fruit, toy, container, bowl, plate separately — do NOT skip any\\
- If an object is inside a bowl/plate, list BOTH the object AND the bowl/plate\\
- Even if you are unsure whether something is relevant, include it as a distractor unless the task description specifically mentions it\\
- DO NOT include: the robot arm, gripper, end effector, or task-relevant objects\\
- DO NOT include: background furniture, table, shelf, wall, floor\\

Respond with ONLY a JSON object, no explanation:
\begin{verbatim}
{
  "relevant_objects": [
    {
      "label": "short name",
      "visual_description": "color, shape, material 
      (e.g. orange-red round plush)"
    }
  ],
  "distractors": [
    {
      "label": "specific name",
      "visual_description": "color, shape, material"
    }
  ]
}
\end{verbatim}

CRITICAL labeling rules:\\
- Use the SAME object names as the task description when possible. If the task says "apple", label it "apple" — NOT "orange fruit" or "red ball".\\
- These may be toy/fake versions of objects. A toy apple is still "apple".\\
- For distractors, be SPECIFIC: use color + object type (e.g. "yellow banana toy", "purple plum toy", "teal bowl") so each distractor can be individually detected.\\
- visual\_description should describe what the object LOOKS LIKE in the image (color, shape, texture) — this helps the detection model find the right object.\\
- Only include objects that are EXPLICITLY part of the task as relevant.\\
- Do NOT include relevant objects in the distractors list.\\
- It is CRITICAL that you list ALL distractors. Missing even one is a failure.\\
Example relevant: "robot arm", "apple"\\
Example distractors: "pineapple toy", "yellow banana toy", "purple plum toy", "orange slice toy", "teal bowl", "gold bowl", "dark blue bowl", "pink bowl"
\end{tcolorbox}  
\vfill

\end{document}

%% file: figs/tamp_countSuccess.tex





\pgfplotstableread[col sep=comma]{data/tamp/countSuccess.csv}\mydata
\begin{tikzpicture}[
  every axis/.style={
    ybar stacked,
    ymin=0,ymax=1.0,
    enlarge x limits=0.2,
    bar width=8mm,
    axis y line*=none,
    axis x line*=bottom,
    tick label style={font=\footnotesize},
    ylabel style={yshift=-6pt},
    width=0.99*\columnwidth,
    height=4cm, 
  },
]
\begin{axis}[ bar shift=-4.5mm,
    ylabel={Success Rate},
    symbolic x coords={blocked2Less, blocked2, combined},
    xticklabels={{Heavy},{Light},{Clutter \& Stack},},
    xticklabel style={align=left},
    xtick=data,
    legend style={at={(0.3,1.05)}, anchor=north west, legend columns=1, draw=none, fill=none},
   ]

\addplot [forget plot, draw=ColorI2, fill=ColorI2] table[x=scene,y=baseline1Percent]{\mydata};
\addplot [forget plot, draw=ColorI1, fill=ColorI1] table[x=scene,y=baseline2Percent]{\mydata};
\addlegendimage{legend image code/.code={
    \fill[ColorI2] (0cm,-0.1cm) rectangle (0.15cm,0.1cm);
    \fill[ColorI1] (0.15cm,-0.1cm) rectangle (0.3cm,0.1cm);
}}
\addlegendentry{Baseline-TAMP Iter 1,2}
\addlegendimage{legend image code/.code={
    \fill[ColorI3] (0cm,-0.1cm) rectangle (0.15cm,0.1cm);
    \fill[Fail]    (0.15cm,-0.1cm) rectangle (0.3cm,0.1cm);
}}
\addlegendentry{\methodname-TAMP Iter 1,2}
\end{axis}
\begin{axis}[bar shift=4.5mm, 
    symbolic x coords={blocked2Less, blocked2, combined},
    xticklabels={{},{},{},},
    xticklabel style={align=left},
    xtick=data,
]
\addplot+[draw=ColorI3, fill=ColorI3] table[x=scene,y=vlm1Percent]{\mydata};
\addplot+[draw=Fail, fill=Fail] table[x=scene,y=vlm2Percent]{\mydata};
\end{axis}
\end{tikzpicture}

%% file: figs/pruningObjectsAblation.tex
\pgfplotstableread[col sep=comma]{data/pushAnything/simObjects.csv}\pruningObjects

\begin{tikzpicture}
\begin{semilogyaxis}[
    xlabel={Number of Objects},
    ylabel={Runtime [s]},
    width=0.4*\columnwidth,
    height=2.2cm, 
    legend style={at={(0.35,1.2)}, 
      anchor=north,legend columns=1, draw=none,fill=none},
    ymajorgrids=true,
    grid style=dashed,
    axis y line*=none,
    axis x line*=bottom,
    scale only axis,
    enlarge x limits=false,
    clip=false,
    ylabel style={yshift=-6pt},
    tick label style={font=\footnotesize},
]

\addplot [error bars/.cd, y dir=both, y explicit,error bar style={ColorBaseline,line width=1.0pt}]
            table [x=NumObjects,y=baseline, y error=bstderr]{\pruningObjects};
\addplot [error bars/.cd, y dir=both, y explicit, error bar style={ColorVLM,line width=1.0pt}]
            table [x=NumObjects,y=vlm, y error=vstderr]{\pruningObjects};

\addplot[color=ColorBaseline,mark=asterisk, line width=3.0pt] table[x=NumObjects,y=baseline]{\pruningObjects};
\addplot[color=ColorVLM,mark=asterisk, line width=3.0pt] table[x=NumObjects,y=vlm]{\pruningObjects};

\legend{,,Baseline-C3, \methodname-C3}

\end{semilogyaxis}
\end{tikzpicture}

%% file: figs/hardware.tex
\pgfplotstableread[col sep=comma]{data/pushAnything/hardwareBoth.csv}\hardwareData

\begin{tikzpicture}
\begin{axis}[
    xbar stacked,
    bar width=3mm,
    xmin=0,  xmax=1.0,
    enlarge y limits=0.8,
    legend style={ at={(0.5, 1.4)},
      anchor=north,legend columns=4, draw=none,},
    xlabel={\% Trials},
    symbolic y coords={pruning, merging},
    yticklabels={{Pruning}, {Merging}},
    yticklabel style={align=center},
    ytick=data,
    axis y line*=none,
    axis x line*=bottom,
    tick label style={font=\footnotesize},
    y tick label style={anchor=east},
    width=0.95*\columnwidth, height=2.7cm, 
    ]
\addplot+[xbar,fill=ColorI1,draw=ColorI1] table[x=one,y=condition]{\hardwareData};
\addplot+[xbar,fill=ColorI2,draw=ColorI2] table[x=two,y=condition]{\hardwareData};
\addplot+[xbar,fill=ColorI3,draw=ColorI3] table[x=three,y=condition]{\hardwareData};
\addplot+[xbar, fill=Fail, draw=Fail] table[x=fail,y=condition]{\hardwareData};

\legend{Iteration 1, Iteration 2, Iteration 3, Fail}
\end{axis}

\end{tikzpicture}

%% file: figs/baselines.tex
\begin{tikzpicture}
\begin{axis}[
    xbar stacked,
    bar width=3.5mm,
    xmin=0, xmax=1.0,
    enlarge y limits=0.1,
    legend style={
        at={(0.5, 1.15)},
        anchor=north,
        legend columns=4,
        draw=none,
    },
    xlabel={\% Trials},
    xtick={0, 0.2, 0.4, 0.6, 0.8, 1.0},
    xticklabels={0, 0.2, 0.4, 0.6, 0.8, 1},
    ytick={0, 1, 2, 3, 4, 5, 6},
    yticklabels={
        {Radius 0.05},
        {Radius 0.1},
        {Radius 0.15},
        {Radius 0.2},
        {Radius 0.25},
        {Nearest 3},
        {VLM-Focus}
    },
    yticklabel style={align=right},
    axis y line*=none,
    axis x line*=bottom,
    tick label style={font=\footnotesize},
    y tick label style={anchor=east},
    width=0.95\columnwidth,
    height=5cm,
]

\addplot+[xbar, fill=ColorI1, draw=ColorI1] coordinates {
    (0.033, 0)
    (0.200, 1)
    (0.233, 2)
    (0.267, 3)
    (0.100, 4)
    (0.267, 5)
    (0.600, 6)
};

\addplot+[xbar, fill=ColorI2, draw=ColorI2] coordinates {
    (0.133, 0)
    (0.200, 1)
    (0.133, 2)
    (0.233, 3)
    (0.033, 4)
    (0.067, 5)
    (0.033, 6)
};

\addplot+[xbar, fill=ColorI3, draw=ColorI3] coordinates {
    (0.133, 0)
    (0.067, 1)
    (0.200, 2)
    (0.033, 3)
    (0.033, 4)
    (0.133, 5)
    (0.133, 6)
};

\addplot+[xbar, fill=Fail, draw=Fail] coordinates {
    (0.700, 0)
    (0.533, 1)
    (0.433, 2)
    (0.467, 3)
    (0.833, 4)
    (0.533, 5)
    (0.233, 6)
};

\legend{Iteration 0, Iteration 1, Iteration 2, Fail}

\end{axis}
\end{tikzpicture}

%% file: figs/vlaSuccess.tex
\pgfplotstableread[row sep=\\,col sep=&]{
    type & BaseClutter & VLAClutter & Base \\
    Model     & 0.5  & 0.69  & 0.85 \\
    }\mydata

\begin{tikzpicture}
\begin{axis}[
    xbar,
    xmin=0,  xmax=1.0,
    enlarge y limits=0.2,
    legend style={at={(0.5,-0.5)},
      anchor=north,legend columns=1, draw=none,},
    xlabel={Success Rate},
    symbolic y coords={Model},
    yticklabels={{Baseline \\without Clutter \vspace{1mm}\\ LENS \\ in Clutter \vspace{1mm}\\ Baseline \\ in Clutter},},
    yticklabel style={align=center},
    ytick=data,
    bar width=6mm,
    axis y line*=none,
    axis x line*=bottom,
    tick label style={font=\footnotesize},
    width=0.9*\columnwidth,
    height=4cm, 
    ]
\addplot [draw=ColorBaseline, fill=ColorBaseline] table[x=BaseClutter,y=type]{\mydata};
\addplot [draw=ColorVLM, fill=ColorVLM] table[x=VLAClutter,y=type]{\mydata};
\addplot [draw=ColorBaselineClutter, fill=ColorBaselineClutter] table[x=Base,y=type]{\mydata};
\end{axis}
\end{tikzpicture}